\documentclass[runningheads]{llncs}
\usepackage{graphicx}
\usepackage{cite}
\usepackage{amsmath,amssymb,amsfonts}

\usepackage{algorithm}
\usepackage{algpseudocode}

\begin{document}

\title{Hierarchical clustering of complex energy systems using pretopology}

%
%
\author{Loup-Noé Lévy\inst{1, 2} \and
Jérémie Bosom\inst{2,3} \and
Guillaume Guerard\inst{4} \and
Soufian Ben Amor\inst{1} \and
Marc Bui\inst{3} \and
Hai Tran\inst{2}}
\authorrunning{L-N. Lévy et al.}
%

\institute{LI-PARAD Laboratory EA 7432, Versailles University, 55 Avenue de Paris, 78035 Versailles, France
\email{\{f\_author. s\_author\}@uvsq.fr} \and Energisme, 88 Avenue du Général Leclerc, 92100 Boulogne-Billancourt France \email{\{f\_author. s\_author\}@energisme.com} \and EPHE, PSL Research University, 4-14 Rue Ferrus, 75014 Paris, France \email{\{f\_author. s\_author\}@ephe.psl.eu} \and De Vinci Research Center, Pole Universitaire Léonard de Vinci, 12 Avenue Léonard de Vinci, 92400 Courbevoie, France \and email{\{f\_author. s\_author\}@devinci.fr}}

\maketitle              
\begin{abstract}
This article attempts answering the following problematic: How to model and classify energy consumption profiles over a large distributed territory to optimize the management of buildings' consumption?

Doing case-by-case in depth auditing of thousands of buildings would require a massive amount of time and money as well as a significant number of qualified people.
Thus, an automated method must be developed to establish a relevant and effective recommendations system.

To answer this problematic, pretopology is used to model the sites' consumption profiles and a multi-criterion hierarchical classification algorithm, using the properties of pretopological space, has been developed in a Python library.

To evaluate the results, three data sets are used: A generated set of dots of various sizes in a 2D space, a generated set of time series and a set of consumption time series of 400 real consumption sites from a French Energy company.

On the point data set, the algorithm is able to identify the clusters of points using their position in space and their size as parameter.
On the generated time series, the algorithm is able to identify the time series clusters using Pearson's correlation with an Adjusted Rand Index (ARI) of 1.

\end{abstract}

\section{Introduction}
\label{sec:introduction}

In 2015 was signed the Paris agreement in which government from all over the world undertook to keep global warming behind a 2$^{\circ}$C increase compared to the temperatures of 1990.
The year of the Cop21, the worldwide buildings sector was responsible for 30\% of global final energy consumption and nearly 28\% of total direct and indirect CO$_2$ emissions.
Yet the energy demand from buildings and building's construction still rises, driven by improved access to energy in developing countries, greater ownership and use of energy-consuming devices and rapid growth in global buildings floor area, at nearly 3\% per year \footnote{http://www.eia.gov/}.
The International Energy Agency's Reference Technology Scenario (RTS), which accounts for existing building energy policies and climate-related commitments, shows that final energy demand in the global buildings sector will increase by 30\% by 2060 without more ambitious efforts to address low-carbon and energy-efficient solutions for buildings and construction.
As a result, buildings-related CO$_2$ emissions would increase by another 10\% by 2060, adding as much as 415 GtCO$_2$ to the atmosphere over the next 40 years – the half of the remaining 2$^{\circ}$C carbon budget and twice what buildings emitted between 1990 and 2016.\footnote{https://www.iea.org/topics/energyefficiency/buildings/}
Yet there are significant opportunities for improvement, as in the United States where 16\% of energy savings could be achieved by reducing performance deficiencies \cite{mills2011building}. Energy actors such as Trusted Third-Party for Energy Measurement and Performance can play a role in identifying the most relevant actions to optimize energy consumption by exploiting the massive energy data now available \cite{bosom2020conception}.

There are many ways to decrease buildings' energy consumption \cite{guerard2017demand}: social programs, incentive programs, new energies, energy efficiency, dynamic pricing, demand-response programs.
But it is challenging to identify precisely what action to take. 

Furthermore, the energy systems are not necessarily buildings.
They can be a building floor or simply a place inside a building.
In consequence, it is more accurate to talk about \textbf{sites} \cite{bosom2020conception}.

The scales of analysis are various both in time (consumption time series are analyzed on a 24h profile as well as on a yearly profile) and space (the studied system can go from one room to a group of buildings across a country).
Because of that, there is no universal performance scale on which to compare a site to another.

Because sites present an important heterogeneity both in intrinsic properties and geographic situation \cite{miller2016screening} only a comparison between similar sites might be meaningful to understand the performance of a new site.
By investigating the works that were effective on a certain site, one can deduce what programs will probably be efficient for sites of similar nature.
Hence, clustering sites based on their characteristics and consumption will enhance their evaluation and the recommendations system.

Therefore the topic of our paper is as following: \textit{How to cluster a large number of heterogeneous sites based on their energy consumption profiles to recommend the most relevant energy optimization solution possible?}

In this article, we will consider that the energy consumption profile encompasses all the physical characteristics of a site as well as the external factors and the consumption data (time series, categorical data and numerical data). The latter is considered as a time series.

Our goal is to study a group of sites to optimize their consumption thanks to recommendations done on similar sites.
This can be assimilated to portfolio analysis.
Portfolio analysis represents a domain in which a large group of buildings, often located in the same geographical area or owned or managed by the same entity, are analyzed for the purpose of managing or optimizing the group as a whole \cite{miller2016screening}.

The key contribution of this paper is to provide a clustering method adapted to portfolio analysis based on a pretopological framework.
- new definitions, properties, and demonstrations
- detailed explanations of the algorithms and their pseudo-codes

Compared to the previous paper \cite{levy2021application} this paper gives greater theoretical understanding of pretopology through added definitions, properties, and demonstration. It demonstrates how the pretopological framework used for the algorithms allows for the clustering of any finite set of items. It also explains the algorithms in greater details as well as presenting the pseudo-code of the algorithms. It also discusses the future work to exploit clustering for energy performance

The paper is structured as follows: the section \ref{sec:State} introduces clustering methods and some relevant examples on energy systems.
The section \ref{sec:Pretopology} presents the pretopology theory and the different types of pretopological spaces.
The section \ref{sec:Hierarchical Clustering Algorithms} explains in details the algorithms developped in the python library with pseudo-code, demonstrating how all finite set of items can be hierarchically clustered.
The section \ref{sec:Model validation and visualization of results} presents the clustering of different types of datasets.
We discuss the results and futur work in the section \ref{sec:Discussion and futur work}
We conclude in the section \ref{sec:Conclusion}.

\section{Literature review}
\label{sec:State}

In this section, we present clustering methods and their application on energy systems.
Clustering is a set of unsupervised machine learning methods that group unlabeled items into clusters.
The journal paper of Iglesia et al. in Energies \cite{iglesias2013analysis} presents a deeper analysis of clustering in energy system. To consult an exhaustive list of clustering algorithms, please read Xu et Al. survey \cite{xu2015comprehensive}.

There are four classes of clustering algorithms. Each of them having pros and cons: density-based clustering, centroid-based clustering, hierarchical clustering, distribution-based clustering.
Let us present each class and their application to portfolio analysis in energy system.

\paragraph{Centroid-based clustering:}

In these methods, a cluster is a set of items such that an item in a cluster is closer to the center of a cluster than to the center of any other cluster.
The center of a cluster is called the centroid, the average of all points in the cluster, or medoid, the most representative point in a cluster.
The well-known centroid-based algorithm is the \textit{K-means} algorithm and its extensions.
The \textit{K-means} algorithm is a powerful tool for clustering, but it requires to determine in advance the number of clusters that the algorithm should find.

Therefore, centroid-based algorithms are sensitive to initial conditions.
Clusters vary in size and density and include outliers (isolated items) from the nearest cluster.
Finally, centroid-based algorithms do not scale with the number of items and dimensions.
In this case, centroid-based algorithms are combined with principal component analysis or spectral analysis to be more efficient.

Regarding portfolio analysis in energy systems, Gao et al. \cite{GAO2014607} compare a multidimensional energy consumption dataset using a \textit{k-means} algorithm.
Freischhacker et al. \cite{FLEISCHHACKER20191092} design a spatial aggregation method, combined with \textit{k-means}, based on block characteristics to reduce reductions due to energy consumption.

\paragraph{Density-based clustering:}
In density-based clustering, a cluster is a set of features distributed in the data space over a contiguous region of high feature density.
Elements located in low density regions are generally considered noise or outliers \cite{kriegel2011density}.
The well-known methods in this class are Density-Based Spatial Clustering of Applications with Noise (\textit{DBSCAN}) and its extensions.

Two parameters influence the formation of clusters: density and accessibility.
Therefore, clusters are distinct according to these parameters.
The main strength of this density-based clustering algorithm is it does not require apriori specification and that it is able to identify noisy data during clustering.
It fails in the case of neck type datasets and does not work well for high dimensionality data.

Regarding portfolio analysis in energy systems, Li et al. \cite{LI2020123115} present a density-based method with particle swarm optimization of building portfolio parameters. Their method predicts the next day's electricity consumption through clustering. Marquant et al. \cite{MARQUANT201873} use a density and load-based algorithm to facilitate large-scale modeling and optimization of urban energy systems.

\paragraph{Hierarchical clustering:}
Hierarchical clustering is most often a procedure whose goal is to transform a proximity matrix into a sequence of hierarchically structured partitions.

The two methods of hierarchical clustering are the bottom-up method (or agglomeration) or the top-down method (or division).
Bottom-up methods start from disjoint classes and place each of the elements in an independent class.
From the proximity matrix, the procedure searches at each step for the two closest classes, merges them, then places them in a second partition.
The process is repeated to build a sequence of nested partitions in which the number of classes decreases as the sequence progresses until a unique class contains all elements.
Top-down methods perform the reverse process.

The key problem of these algorithms is to define the criterion for grouping or aggregating two classes, i.e. a distance measure.
Sites are defined as complex systems:\cite{ahat2013smart,Bosom2018multi,7298002,bosom2020conception}.
They are defined with numerical and categorical data as well as time series. For this reason calculating a distance between two elements is challenging and does not allow to use every feature of the site in a relevant way.
Another drawback is the difficulty of identifying a precise number of clusters, especially in a large data set.

Regarding portfolio analysis in energy systems, Wang et al. \cite{WANG2020117195} analyze the spatial disparity of final energy consumption in China through hierarchical clustering and spatial autocorrelation. Li et al. \cite{LI2019735} implement a strategy based on agglomerative hierarchical clustering to identify typical daily electricity usage patterns.

\paragraph{Distribution-based clustering:} Application to large spatial databases requires from clustering algorithms to have no or minimal input parameters and arbitrarily shaped clusters.
Distribution-based clustering produces clusters that assume concisely defined mathematical models underlying the items, a relatively plausible assumption for some item distributions.

Most of the time, the mathematical models are based on the Gaussian, multinomial, or multivariate normal distribution.
Clusters are considered fuzzy, which means that an item can be found in several clusters at a defined percentage. The best known algorithm is the Expectation-Maximization (EM) clustering with Gaussian mixture models (GMM).
Thus, the GMM algorithm provides two parameters to describe the shape of the clusters: the mean and the standard deviation.
The main drawback of these algorithms is that they cannot work on categorical dimensions.

Regarding portfolio analysis in energy systems, Lu et al. \cite{LU201949} use GMM clustering for the identification of heating load patterns. Habib et al. \cite{habib2015outliers} provide EM clustering to detect outliers in the energy building portfolio.

\paragraph{Conclusion about clustering methods:}
None of the methods described above can answer the specificities of the studied system, either because they require the definition of a distance between the items, or because they cannot return the hierarchical clustering necessary to apprehend the different scales of a complex system.

\paragraph{Relevance of pretopology-based clustering:}

A pretopological space is defined by a relationship between a set of items and a larger set of items. It is therefore suitable for creating a hierarchical structure. It is based on the concept of abstract space. In such a space, the nature of the element is not relevant, it is rather the relations and properties linking the elements together that are important. This allows us to manipulate heterogeneous and complex elements such as our sites.
Therefore, pretopology can be considered as a mathematical tool to model the concept of proximity for complex systems \cite{Auray2009Pretopology}.
Pretopology is therefore the approach chosen to build our hierarchical clustering.

\section{Pretopology}
\label{sec:Pretopology}

In this section we will explain the key concepts and definition of pretopology, such as pretopological space and pseudo-closure.
We won't go into detail on the origins of pretopology but it is important to understand that the concept of pretopological space is obtained by weakening the hypothesis of the topological spaces. It allows the modeling of discrete structures unlike topology \cite{Auray2009Pretopology}.

\subsection{Pretopological space}
\subsubsection{Central definitions and propositions}
\begin{definition} A pseudoclosure function $a : \wp(U) \to \wp(U)$ on a set $U$, is a function such that:
\begin{itemize}
    \item $a(\emptyset) = \emptyset$
    \item $\forall A \mid A \subseteq U : A \subseteq a(A)$
\end{itemize}
where $\wp(U)$ is the power set of U
\end{definition}

\begin{figure}
  \centering
  \includegraphics[width=180pt]{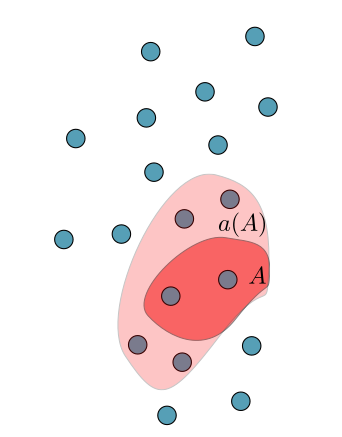}
  \caption{Example of a pseudoclosure function\cite{laborde2019Pretopology}}
  \label{fig:pseudoclosure}
\end{figure}

\begin{definition} A tuple $(U, a(.))$, where $U$ is a set of elements and $a(.)$ is a pseudoclosure function on $U$, constitutes a pretopological space.
\end{definition}

We note that a pretopological space is defined by establishing a relation between any set of elements and a bigger set. This is interesting in the construction of a hierarchy.
The previous definition determines the most general pretopological space. By asking
the function to fulfill some additional conditions we get more specific pretopological
spaces:

\begin{definition} If $\forall$ $A, B$ $\vert$ $A \subseteq U$, $B \subseteq U$ : $A \subseteq B$ $\implies$ $a(A) \subseteq a(B)$, then we get a pretopological space of type $V$. This property is called isotony.
\end{definition}

\begin{definition} If $\forall$ $A, B$ $\vert$ $A \subseteq U$, $B \subseteq U$ : $a(A \cup B) = a(A) \cup a(B)$, then we get a pretopological space of type $V_D$.
\end{definition}

\begin{definition} If $\forall$ $A$ $\vert$ $A \subseteq U$ : $a(A) = \bigcup_{x \in A} a(x)$ then we get a pretopological space of type $V_S$.
\end{definition}

Given any pretopological space $(U, a(.))$, we can ask ourselves the question of what becomes of the concepts of closure classically defined in topology. In fact, the definition remains the same in pretopology\cite{le2007classification}.

\begin{definition} A part $F$ of $U$ will be a closure of $U$ if and only if $a(F) = F$
\end{definition}

\begin{proposition} In a pretopological space of type $V$, the intersection of closures is a closure.
\end{proposition}

\begin{proposition} In a pretopological $V-type$ space, the closure and opening of any part of U still exists.
\end{proposition}

\begin{proposition} In a pretopological space of type $V$, the closure of a part $A$ of $U$ is the smallest closure containing $A$. Denoted F(A).
\end{proposition}

\begin{proposition} In a pretopological space of type $V$, every set has a closure.
The proof can be found in \cite{belmandt1993manuel}.
\end{proposition}

\begin{figure}[ht]
  \centering
  \includegraphics[width=300pt]{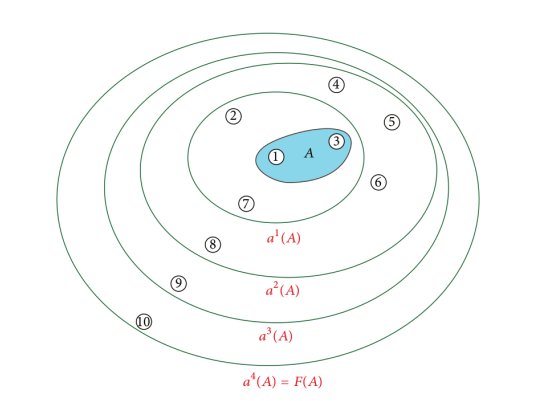}
  \caption{Closure of set $A$\cite{laborde2019Pretopology}}
  \label{fig:closure}
\end{figure}

In a pretopological space of type $V$ we can find the closure by repeatedly applying the pseudoclosure operator to the set and its subsequent images until it stops expanding.
We can see an example of this in figure \ref{fig:closure} \cite{laborde2019Pretopology}.

\begin{figure}[ht]
  \centering
  \includegraphics[width=270pt]{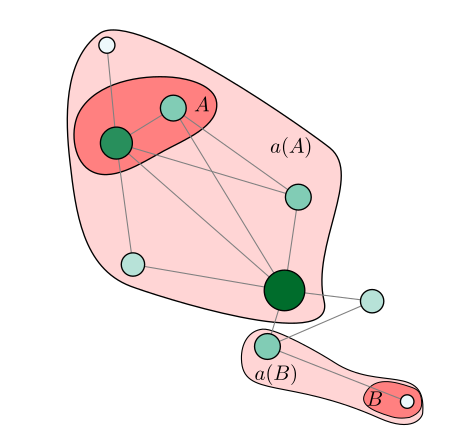}
  \caption{Pseudoclosure function on a graph\cite{laborde2019Pretopology}}
  \label{fig:pseudoclosure_graph}
\end{figure}

If we have a pretopological space of type $V_D$ and $\forall A$ $\vert$ $A \subseteq U$ : $a(A) =
a(a(A))$, then we get a topology. The pseudoclosure function here is said to be idempotent \cite{laborde2019Pretopology}. It’s clear that in a finite space, $V_S$ = $V_D$ \cite{belmandt1993manuel}. Also, in pretopological spaces of type $V_D$ the pseudoclosure of a set is completely defined by the pseudoclosures of its singletons. So if the space is also finite, we could draw an edge from an element to every element of its pseudoclosure, and the pseudoclosure would be equivalent to a particular graph. Figure \ref{fig:pseudoclosure_graph} shows the relation between the two. This demonstrates that pretopology is also a generalization of graph theory \cite{laborde2019Pretopology}.

There is a second way of characterizing pretopologies of type $V$ and $V_D$ . To understand
it we need to give a few more definitions first:
\begin{definition} We say that a set $\mathcal{F}$ of $\wp(\wp(U))$ is a prefilter over $U$, if:
\begin{equation}
\forall F \in \mathcal{F},  \forall H \in \wp(U), F \subset H \Longrightarrow H \in \mathcal{F}
\end{equation}

\end{definition}

\begin{definition} We say that a set $\mathcal{F}$ of $\wp(\wp(U))$ is a filter over $U$, if it is a prefilter stable under finite intersection, i.e.
\begin{equation}
\forall F \in \mathcal{F}, \forall G \in \mathcal{F}, F \cap G \in \mathcal{F}
\end{equation}
\end{definition}

\begin{figure}[ht]
  \centering
  \includegraphics[width=300pt]{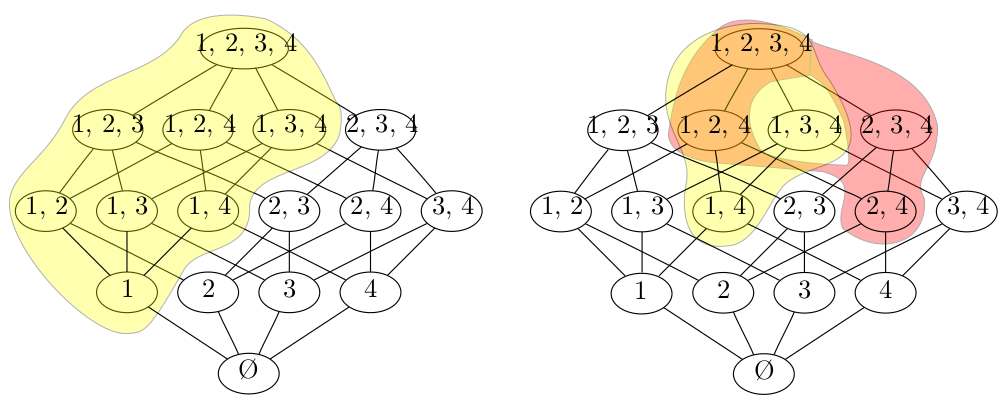}
  \caption{Filters vs Prefilters\cite{laborde2019Pretopology}}
  \label{fig:filtervsprefilter}
\end{figure}

In other words, and restricting ourselves to a finite space, a filter is the family of all
supersets of a set $\mathcal{B}$, while a prefilter is the family of supersets of every member $B_i$ of a family of sets $\mathcal{B}$. The family of sets $\mathcal{B}$ is called the basis of the prefilter. 
We can see in figure \ref{fig:filtervsprefilter} an example of a filter and a prefilter with basis $B$ = {{1, 4}, {2, 4}}\cite{laborde2019Pretopology}.

Now, if we have a set $U$, and for every $x \in U$ we have a prefilter $V (x)$ such that every member of $V (x)$ contains the element $x$, we can define a pseudoclosure function in the following way:
\begin{equation}
\forall A \subseteq U, a(A) = \{x \in U \vert \forall V \in V (x), V \cap A, \emptyset\}
\end{equation}

\begin{figure}[ht]
  \centering
  \includegraphics[width=\textwidth]{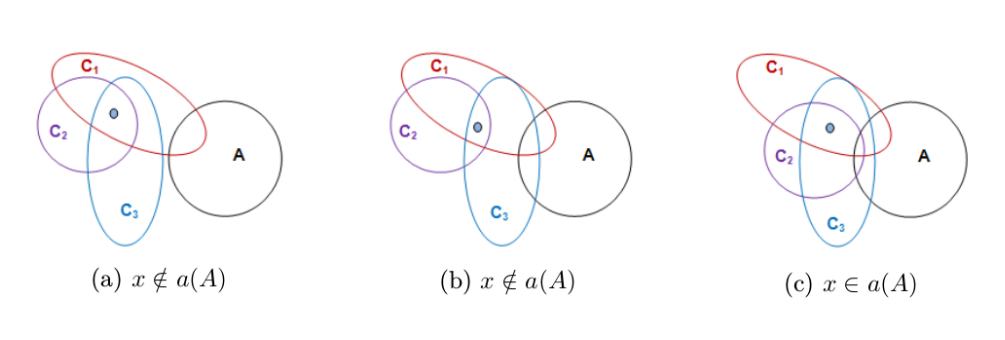}
  \caption{Neighborhood definition of a pretopology\cite{laborde2019Pretopology}}
  \label{fig:neighborhood}
\end{figure}

We call the prefilter $V (x)$ the family of neighborhoods of $x$, and each set in the family is called a neighborhood of $x$. Figure \ref{fig:neighborhood} shows a graphical representation of this definition of the pseudoclosure.
On the other hand, if we have a pseudoclosure function $a(.)$ in a pretopological space of type $V$, the family of sets given by:
\begin{equation}
V (x) = \{V \subset U \vert x \in i(V )\}
\end{equation}
where $i(A) = a(A^c)^c$, is a prefilter.
The following proposition shows that we can go from one definition to the other interchangeably\cite{belmandt1993manuel}:

\begin{proposition} No two families of prefilters $\{V (x_i) \vert x_i \in U\}$ define the same pseudoclosure function $a(.)$, and no two pseudoclosure functions define the same family of prefilters $\{V (x_i) \vert x_i \in U\}$.
\end{proposition}

\section{Hierarchical Clustering Algorithms}
\label{sec:Hierarchical Clustering Algorithms}
This section describes the algorithms developped in a Python library used for the construction of a closure and to build a hierarchical clustering of sites.
This algorithm, whose pseudo-code is given in the source code \ref{listing:appendix:pretopology:structuralAnalysis}, is organized in four phases:
\begin{itemize}
    \item Determine a family of elementary subsets called seeds.
    \item Construct the closures of the seeds by iterative application of the pseudoclosure function.
    \item Construct the adjacency matrix representing the relations between all the identified subsets (even the intermediate ones).
    \item Establish the quasi-hierarchy by applying the associated algorithm on the adjacency matrix.
\end{itemize}

\begin{algorithm}
\caption{\textbf{QuasistructuralAnalysis}: Algorithm for the analysis of the quasi-hierarchy of a pretopological space, based on the work of \cite{laborde2019Pretopology}}
\begin{algorithmic}
\Require $((U,\, a(.)),\, d,\, seedFunc(.),\, th_{qh})$
\Ensure $QF_{qh},\, quasiHierarchy$
\State $seedList \leftarrow ElementaryQuasiclosures((U,\, a),\, d,\, seedFunc)$
\State $QF_e \leftarrow ElementaryClosedSubsets((U,\, a),\, seedList)$
\State $Adj_{qh} \leftarrow ExtractAdjencyQuasihierarchy(QF_e)$
\State $QF_{qh},\, quasiHierarchy \leftarrow ExtractQuasihierarchy(QF_e,\, Adj_{qh},\, th_{qh})$
\end{algorithmic}
\label{listing:appendix:pretopology:structuralAnalysis}
\end{algorithm}

Several methods are possible to determine the seeds.
Therefore, the algorithm is influenced by the following two hyperparameters:
\begin{itemize}
    \item the $seedFunc(.)$ function which determines, for an element, a set of close elements which will constitute a seed,
    \item the degree $d$ to specify the size of the seeds.
\end{itemize}

The algorithm takes an additional hyperparameter, required by the \textit{ExtractQuasihierarchy} algorithm
in order to establish the quasi-hierarchy: $th_{qh}$, which corresponds to the threshold beyond which it is estimated that two elements are close.
This number is generally between $0$ and $1$.

We now detail each phase of the algorithm.

\paragraph{Calculation of a family of elementary sets or seeds}
\label{paragraph:appendix:codeExamples:algos:pretopology:algorithmConception:algoFamilyElementarySeed}

The aim here is to determine elementary subsets of size $d$ called seeds thanks to the function $seedFunc(.)$ whose role is to find the $d$ needed neighbors.
To do so, we iterate on all the points of the set $U$ associated to the pretopological space $p$.
The pseudo-code of the resulting algorithm (named $ElementaryQuasiclosures$) is presented in the source code \ref{listing:appendix:pretopology:elementaryQuasiclosures}.

\begin{algorithm}
\caption{\textbf{ElementaryQuasiclosures}: Construction of the seeds by applying the function $seedFunc(.)$ on all the elements of the set $U$, based on the work of \cite{laborde2019Pretopology}}
\begin{algorithmic}
\Require $((U,\, a(.)),\, degree,\, seedFunc(.))$
\Ensure $seedList$
\State $seedList \leftarrow list()$
\ForAll{$x \in U$}
\State $seedList \leftarrow list()$
\State $seedList.append(seed)$
\EndFor
\end{algorithmic}
\label{listing:appendix:pretopology:elementaryQuasiclosures}
\end{algorithm}

\begin{algorithm}
\caption{\textbf{FindNeighbors}: Determine the $d$ neighbors of $firstNode$ using the $seedFunc(.)$ function, based on the work of \cite{laborde2019Pretopology}}
\begin{algorithmic}
\Require $(firstNode,\, d,\, seedFunc(.))$
\Ensure $path$
\State $path \leftarrow list()$
\State $lastTreatedNode \leftarrow firstNode$
\ForAll{$ i \in range(d)$}
\State $newNode \leftarrow seedFunc(lastTreatedNode)$
\State $path.append(newNode)$
\State $lastTreatedNode \leftarrow newNode$
\EndFor
\end{algorithmic}
\label{listing:appendix:pretopology:findNeighbors}
\end{algorithm}

The algorithm \ref{listing:appendix:pretopology:elementaryQuasiclosures} uses the function $FindNeighbors$ whose peudo-code is given in the source code \ref{listing:appendix:pretopology:findNeighbors}.
The latter takes as parameters an element of $U$, the number of neighbors sought $d$ and the function determining the nearest neighbors $seedFunc(.)$.
The $seedFunc(.)$ function usually takes as its value one of the following two functions:
\begin{itemize}
    \item $ClosestNode(node)$ which identifies the closest nodes to an element. It is used in cases where a distance can be calculated, for example in the case where the studied relations are quantifiable.
    \item $RandomNeighbor(node)$ randomly browses the neighboring nodes. Its use is preferred when the relations are not quantifiable, for example in the case of values describing categories.
\end{itemize}

\paragraph{Construction of subsets by applying pseudoclosure}
\label{paragraph:appendix:codeExamples:algos:pretopology:algorithmConception:subsetCreation}

To construct the subsets that will then be organized by the pseudo-hierarchy algorithm, \textit{ElementaryClosedSubsets} uses the seed list $seedList$ computed previously by \textit{ElementaryQuasiclosures}.
For each of the seeds in $seedList$, the membership function is applied iteratively until the pseudo-closure no longer gives bigger sets.

The intermediate and final subsets are stored in a list of unique element lists (\textit{list} of \textit{set}) named $QF_{tmp}$ so that we don't have to reapply the membership later on the same sets.
$QF_{tmp}$ indexes the subsets according to the number of elements they contain.
Since the membership of a set is always greater than or equal to its size, such indexing ensures that all elements are processed once and only once.

The list $QF_e$, constructed from the lists in $QF_{tmp}$, is then returned.
The associated pseudo-code is presented in the source code \ref{listing:appendix:pretopology:elementaryClosedSubsets}.

\begin{algorithm}
\caption{\textbf{ElementaryClosedSubsets}: Computes the set of subsets by iterative application of the pseudo-closure function, algorithm inspired from \cite{laborde2019Pretopology}}
\begin{algorithmic}
\Require $((U,\, a(.)),\, seedList)$
\Ensure $QF_e$
\State $QF_{tmp}$ a list of $Size(U)$ sets
\ForAll{$seed \in seedList$}
\State $QF_{tmp}[Size(seed)].append(seed)$
\EndFor
\ForAll{$i \in range(1,\, Size(U) + 1)$}
\ForAll{$s \in QF_{tmp}[i]$}
\State $pseudoclosure \leftarrow a(s)$
\If{$lastTreatedNode \leftarrow newNode$}
\State $QF_{tmp}[Size(pseudoclosure)].append(pseudoclosure)$
\EndIf
\EndFor
\EndFor
\State $QF_e \leftarrow list()$
\ForAll{$i \in range(Size(QF_{tmp}))$}
\State $QF_e.extend(QF_{tmp}[i])$
\EndFor
\end{algorithmic}
\label{listing:appendix:pretopology:elementaryClosedSubsets}
\end{algorithm}

\paragraph{Construction of the adjacency matrix}
\label{paragraph:appendix:codeExamples:algos:pretopology:algorithmConception:adjencyMatrix}

The objective of this algorithm is to establish the hierarchical relations between the graphs of $QF_e$ identified by $ElementaryClosedSubsets$.
These relationships, between all $QF_e$ sets, are represented as an adjacency matrix $Adj_{qh}$.

In a space of type $V$, two distinct closed elementary subsets $F_x$ and $F_y$ of $QF_e$ :
\begin{itemize}
    \item are either disjoint then $F_x \cap F_y = \emptyset$,
    \item either contain a nonzero intersection such that $\forall; z \in F_x \cap F_y,\, F_z \subset F_x \cap F_y$, where $F_z$ is the closure of $z$.
\end{itemize}

Thus, if two subsets $F_x$ and $F_y$ overlap without one of them being contained in the other ($F_x \cap F_y \neq \emptyset$, $F_x \not\subset F_y$ and $F_y \not\subset F_x$), we know that a smaller set $F_z$ contained in $F_x \cap F_y$ exists.
The resulting hierarchical graph must therefore connect $F_x$ and $F_y$ as parents of $F_z$.

\begin{figure}[ht]
  \centering
  \includegraphics[width=\columnwidth]{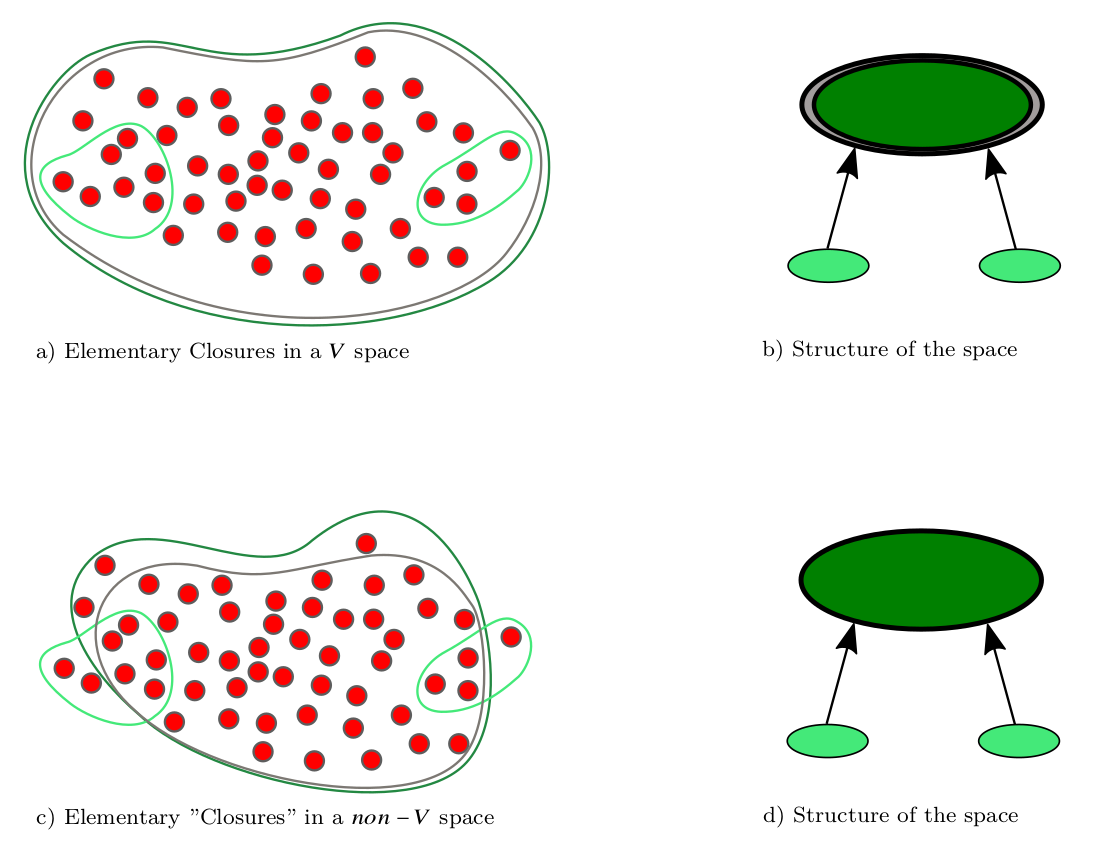}
  \caption{Construction of a quasi-hierarchy in a pretopological space of type $V$, according to the method of \cite{largeron2002pretopological}, and of type non-$V$, according to the method of \cite{laborde2019Pretopology}, figures by this later author}
  \label{fig:hierarchy}
\end{figure}

However, as we mentioned, the Laborde's algorithm \cite{laborde2019Pretopology}  is intended to be applicable to non-$V$ spaces as well.
In such pretopological spaces, there is no guarantee that an element of $F_x \cap F_y$ will not grow beyond this intersection.
This is illustrated in Figure \ref{fig:hierarchy}.
Furthermore, in the case of $d - n$ elementary sets, where $n$ is the cardinality of $U$ and $d$ is the degree applied for creating the seeds, it is possible that none of the seeds are contained in the intersection.
Thus, it is possible that no obvious structure emerges from the collection of quasi-closures.

To solve this problem, Laborde et Al. \cite{laborde2019Pretopology} generalizes the type of hierarchy constructed from quasi-closures so as to satisfy the following constraints:
\begin{itemize}
    \item Two subsets should be connected only if their intersection is nonzero ($F_x \cap F_y$),
    \item The larger the cardinality of the intersection $F_x \cap F_y$ is compared to that of $F_x$, the stronger the relation of $F_x$ to $F_y$ is,
    \item The larger the cardinality of the subset $F_y$ compared to that of $F_x$, the less necessary it is that $F_x \cap F_y$ is large for the relation from $F_x$ to $F_y$ to be strong. In other words, a very large set will attract smaller sets even if their intersection is not very large.
\end{itemize}

The algorithm presented in the source code \ref{listing:appendix:pretopology:quasihierarchyAdjencyMatrix} implements this logic.
It quantifies the relations between each pair of $QF_e$ whose intersection is not empty and then returns the resulting matrix.

\begin{algorithm}
\caption{\textbf{ExtractAdjencyQuasihierarchy}: Construction of the adjacency matrix for the quasi-hierarchy, algorithm inspired from \cite{laborde2019Pretopology}}
\begin{algorithmic}
\Require $(QF_e)$
\Ensure $Adj_{qh}$
\State $Adj_{qh} \leftarrow SquaredMatrixZeros(size(QF_e))$
\ForAll{$F,\, G \in QF_e$}
\State $F has G \leftarrow Size(F \cap G)/Size(G)$
\State $G has F \leftarrow Size(F \cap G) / Size(F)$
\State $F bigger G \leftarrow Size(F) / Size(G)$
\State $G bigger F \leftarrow Size(G) / Size(F)$
\State $Adj_{qh}[Index(G),Index(F)] = G bigger F * G has F$
\State $Adj_{qh}[Index(F),Index(G)] = F bigger G * F has G$
\EndFor
\end{algorithmic}
\label{listing:appendix:pretopology:quasihierarchyAdjencyMatrix}
\end{algorithm}

\paragraph{Construction of the quasi-hierarchy}
\label{paragraph:appendix:codeExamples:algos:pretopology:algorithmConception:subsetHierarchyCreation}

The quasi-hierarchy is built from the adjacency matrix by checking if the relations computed by \textit{ExtractAdjencyQuasihierarchy} exceed the threshold $th_{qh}$.
The new adjacency matrix thus obtained defines the quasihierarchy returned by the algorithm.
The algorithm also returns the final list $QF_{qh}$ of identified subsets for the set $U$.
$QF_{qh}$ corresponds to the list $QF_e$ updated following the potential addition of new subsets during the construction of the quasi-hierarchy.

The quasi-hierarchy is established by applying the following rules on the values of $Adj_{qh}$:\begin{itemize}
    \item A link between two subsets is established in the quasi-hierarchy if their relationship exceeds the threshold $th_{qh}$,
    \item Two subsets that have strong mutual relations (exceeding the threshold $th_{qh}$) are considered equivalent. They are subject to a subsidiary treatment improving the resulting quasi-hierarchy.
    \item The resulting quasiclosures with the respective links determine the quasihierarchy.
\end{itemize}

Laborde et Al. \cite{laborde2019Pretopology} treats the case of equivalent sets by keeping the largest set and deleting the other.
If the sets are of the same size then one of them is chosen randomly.

\section{Model validation and visualization of results}
\label{sec:Model validation and visualization of results}

\paragraph{Validation tool:}
To evaluate the pretopological hierarchical clustering, we also provide a set of tools to validate the model and show the results.

This program is developed to create a point dataset with the following parameters:
\begin{itemize}
\item[-] the number of groups of dense items;
\item[-] the number of items of each group;
\item[-] the spatial dispersion of each group;
\item[-] the position of each group.
\end{itemize}

The size of an item is added as a second parameter, to evaluate multi-criteria clustering.
Groups with different item size can be produced with the following parameters:
\begin{itemize}
\item[-] the number of groups;
\item[-] the number of items of each group;
\item[-] the range of sizes of each group.
\end{itemize}
This program allows us to evaluate our method in different types of situations and to easily make adjustments or corrections.
\begin{figure*}[ht]
    \centering
    \includegraphics[width=\textwidth]{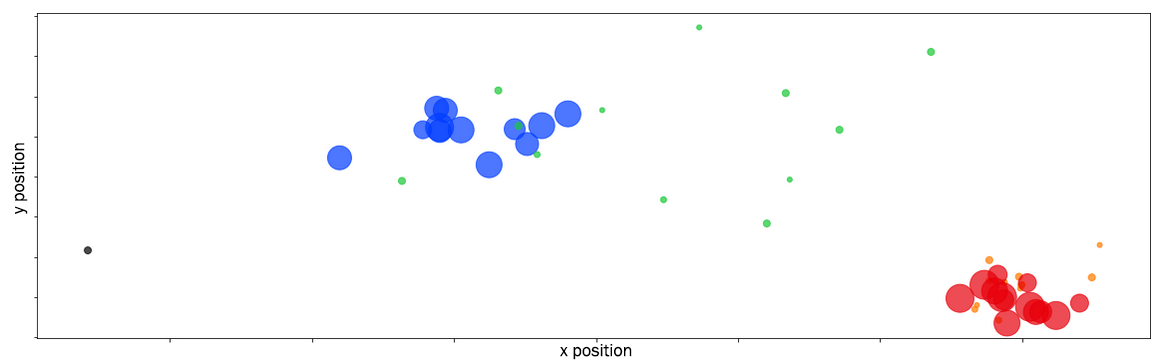}
    \caption{The four clusters determined by our algorithm using both size and position as parameters, on a 2D disks dataset \cite{levy2021application}.}
    \label{fig:postion_size}
\end{figure*}

\paragraph{Visualization tool:}
The program colors each of the largest sets determined by our algorithm with a single color to make the clusters apparent.
The validation tool is tested with two groups of large and small elements and a two-dimensional position. The elements are shown in figure \ref{fig:postion_size}.
In this example, four clusters were determined: blue, green, orange and red.
The black dot at the far left of the figure \ref{fig:postion_size} is an element identified as an outlier by the algorithms.
For example, the red and orange elements are close to each other but separated into two clusters due to their different sizes, and the orange and green dots are similar in size but divided into two sets due to their different positions.

The program also displays the hierarchical classification consisting of the seeds, intermediate sets and final clusters.
The hierarchical classification is displayed as a tree in which each set is identified by a number and is represented by a node.

\begin{figure}[ht]
    \centering
    \includegraphics[width=1.02\columnwidth]{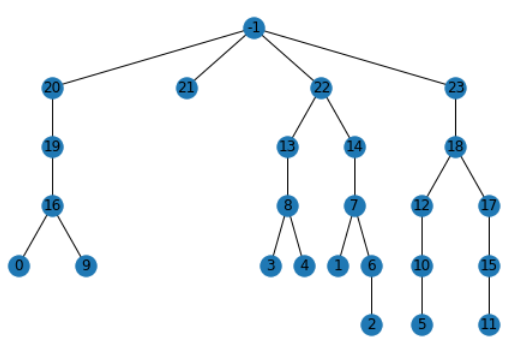}
    \caption{A tree representing the pseudohierarchy relation between each intermediate set from the seed to the cluster \cite{levy2021application}.}
    \label{fig:tree}
\end{figure}

\begin{figure}[ht]
    \centering
    \includegraphics[width=\columnwidth]{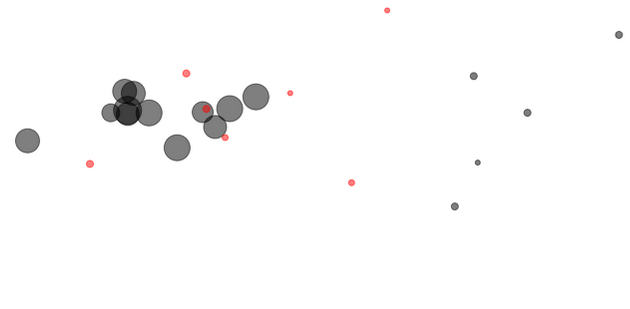}
    \caption{The subset 14 in red representing a subgroup of the green clusters (subset 22) in figure \ref{fig:postion_size} \cite{levy2021application}.}
    \label{fig:14}
\end{figure}

For example, the hierarchy shown in Figure \ref{fig:tree} shows the relationships between the sets determined by our algorithm applied to the dataset displayed in Figure \ref{fig:postion_size}.
Only the sets with more than two elements are shown on this tree.
We can recognize the four clusters that have been colored on figure \ref{fig:postion_size}, they are labeled 20, 21, 22 and 23.
Figure \ref{fig:14} displays cluster 14 which is a child of cluster 21 (colored in green) in the hierarchical clustering.
This hierarchy identifies large clusters of relatively similar items and provides more detail about small clusters of very similar items.

\subsection{Results on different datasets}

\subsection{Benchmark dataset}

\begin{figure}[ht]
    \centering
    \includegraphics[width=\columnwidth]{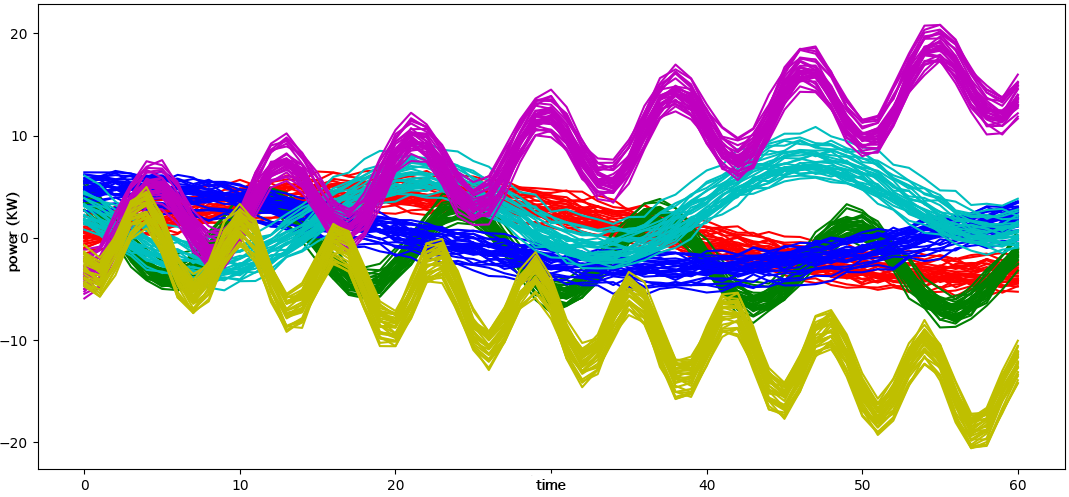}
    \caption{The clusters identified by our algorithm\cite{levy2021application}.}
    \label{fig:ground_truth}
\end{figure}

Since the main data we have from the sites are time series of power consumption, we needed to test, visualize, and evaluate the clustering of a time series set. This section presents this test set and the results of our algorithm.
The test set created, consisting of six clusters, is shown in Figure \ref{fig:ground_truth}.
Each cluster is composed of 30 time series of 60 points.

The similarity measure used to establish the value between two items is the Pearson's coefficient.
The Pearson correlation coefficient measures the linear relationship between each pair of items, which in this case are time series.

Our program colored the time series based on the clusters it had identified (see figure \ref{fig:ground_truth}).

\subsection{Results analysis on benchmark dataset}
\begin{figure}[ht]
    \centering
    \includegraphics[width=0.8\columnwidth]{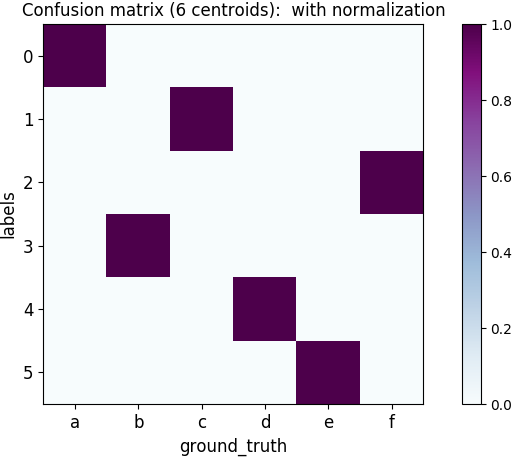}
    \caption{Confusion matrix of the clusterization\cite{levy2021application}.}
    \label{fig:confusionmatrix}
\end{figure}

The program identified exactly the same clusters as the ground truth given by the benchmark.
To evaluate the validity of the clusters determined by the algorithm, our metric is the Adjusted Rand score, also called Adjusted Rand Index (ARI).
As we have perfectly identified the clusters, the ARI of our clustering is $1$.
Figure \ref{fig:confusionmatrix} shows the confusion matrix between the cluster found by our method and the ground truth given by the benchmark.

Further experiments will be conducted in a future contribution.

\subsection{Real dataset}

This dataset is built from Enedis (the French electricity network manager) consumption time series for 400 sites over one year.
It is resampled with a time step of half an hour, a day, a week and a month.
The proximity between Enedis delivery points is evaluated on each resampled time series, each resampled time series corresponding to a characteristic of a site.
After the Enedis dataset is constructed, the algorithm described in section \ref{sec:Pretopology} is applied on the time series.

\subsection{Result Analysis on real dataset}
\begin{figure}[ht]
    \centering
    \includegraphics[width=\columnwidth]{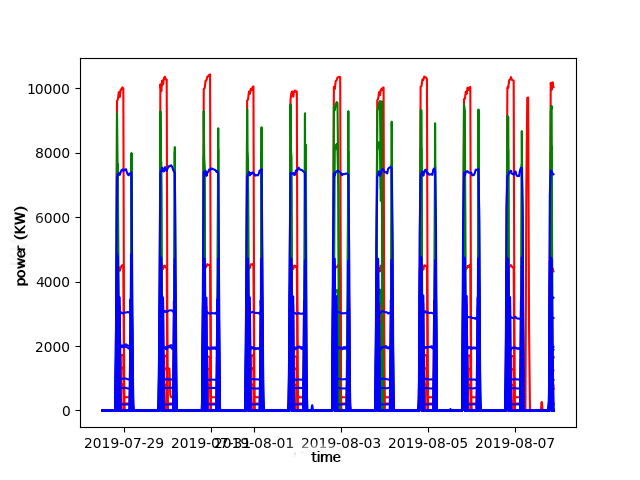}
    \caption{Clustering of the Enedis time series\cite{levy2021application}}
    \label{fig:enedisclassification}
\end{figure}

Figure \ref{fig:enedisclassification}, displays the grouping of 50 Enedis time series representing all the clusters.
Three clusters have been identified, in the green cluster there are two peaks per day, one in the morning, one in the evening, in the red clusters there is a single peak per day that lasts half the day,  and in the blue cluster the consumption is constant during the day.

The algorithm identified relevant clusters in the sense that each items shares a characteristic with items in its cluster that it does not share with items in a different cluster.

\section{Discussion and future work}
\label{sec:Discussion and futur work}
The results we have shown on a real dataset are preliminary.
To fully exhibit the potential of this algorithm, the clustering will have to be applied to a richer data set.
This data set should include relevant features extracted from the consumption time series as well as physical characteristics of the buildings (such as the site's floor area, the type of heating, the insulation material, etc.).
Correlation between the sites consumption and meteorological environment will also be a feature used for future works.
By taking all these elements into account, the relevance of the clusters identified will be greatly improved. 

There are two possibilities regarding the identified building clusters: 

\begin{itemize}
    \item The clusters correspond to an already defined classification i.e. the clusters can be compared to ground truth.
For example, the clusters identified might correspond to the usage of the sites.
In this case, we will implement semi-supervised learning and by using Machine Learning to tune the hyper-parameters of the algorithm we will optimize the ARI index of our clustering.
    \item The clusters do not correspond to any known classification of buildings. In that case, we will have to apply knowledge extraction methods as well as energy experts' insight to give meaning to the newly found taxonomy of buildings.
\end{itemize}

Because the energy performance key indicators are not the same depending on the type of building \cite{boemi2016indicators}, the insight given on building types will enhance energy performance evaluation and recommendation. 

\section{Conclusion}
\label{sec:Conclusion}

Building energy performance is a major challenge of the 21st century because of its important impact on climate change.
Allowed by the growth of energy data, clustering of building based on consumption profile is a promising solution to efficiently identify the most relevant action to take for such complex energy systems.
After a presentation of the state of art of the clustering methods, we propose a novel approach based on pretopology.
The presented framework using pretopology allows for the multi-criteria hierarchical clustering of any finite set of items.
Having a hierarchical structure gives insight into the similarities between building at different scales and therefore should provides a more refined understanding of the families and subfamilies of consumption profiles.
The algorithm developed in Python for the construction of a Hierarchical Clustering of sites exploits this framework.
The validation and visualization tools developed to test our algorithm allowed us to demonstrate visually and through ARI the relevance of the method on generated datasets as well as on real consumption time series dataset.
The results demonstrate the potential of this solution for hierarchical clustering of heterogeneous systems.

\section*{Acknowledgements}
This paper is the result of research conducted at the energy data management company \textit{Energisme}. We thank \textit{Energisme} for the resources that have been made available to us and Julio Laborde for his assistance with the conception of our pretopological hierarchical algorithm library.

\bibliographystyle{splncs04.bst}
\bibliography{mybibliography}

@article{mills2011building,
title = {Building commissioning: a golden opportunity for reducing energy costs and greenhouse gas emissions in the United States},
author = {Evan {Mills}},
year = {2011},
journal = {Energy Efficiency},
page = {145–173}
}

@inproceedings{levy2021application,
  title={Application of Pretopological Hierarchical Clustering for Buildings Portfolio.},
  author={Levy, Loup-No{\'e} and Bosom, J{\'e}r{\'e}mie and Gu{\'e}rard, Guillaume and Amor, Soufian Ben and Bui, Marc and Tran, Ha{\"\i}},
  booktitle={SMARTGREENS},
  pages={228--235},
  year={2021}
}

@incollection{boemi2016indicators,
  title={Indicators for buildings’ energy performance},
  author={Boemi, Sofia-Natalia and Tziogas, Charalampos},
  booktitle={Energy Performance of Buildings},
  pages={79--93},
  year={2016},
  publisher={Springer}
}

@phdthesis{bosom2020conception,
  title={Conception de microservices intelligents pour la supervision de syst{\`e}mes sociotechniques: application aux syst{\`e}mes {\'e}nerg{\'e}tiques},
  author={Bosom, Jérémie},
  year={2020},
  school={Universit{\'e} Paris sciences et lettres}
}

@inproceedings{guerard2017demand,
  title={Demand-Response: Let the Devices Take our Decisions.},
  author={Guerard, Guillaume and Pichon, Bastien and Nehai, Zeinab},
  booktitle={SMARTGREENS},
  pages={119--126},
  year={2017}
}

@PhDThesis{miller2016screening,
title = {Screening Meter Data: Characterization of Temporal Energy Data from Large Groups of Non-Residential Buildings},
author = {C. {Miller}},
school = {ETH Zurich},
year = {2016}
}

@article{iglesias2013analysis,
  title={Analysis of similarity measures in times series clustering for the discovery of building energy patterns},
  author={Iglesias, F{\'e}lix and Kastner, Wolfgang},
  journal={Energies},
  volume={6},
  number={2},
  pages={579--597},
  year={2013},
  publisher={Multidisciplinary Digital Publishing Institute}
}

@article{GAO2014607,
title = "A new methodology for building energy performance benchmarking: An approach based on intelligent clustering algorithm",
journal = "Energy and Buildings",
volume = "84",
pages = "607 - 616",
year = "2014",
issn = "0378-7788",
doi = "https://doi.org/10.1016/j.enbuild.2014.08.030",
url = "http://www.sciencedirect.com/science/article/pii/S0378778814006720",
author = "Xuefeng Gao and Ali Malkawi"
}

@article{xu2015comprehensive,
  title={A comprehensive survey of clustering algorithms},
  author={Xu, Dongkuan and Tian, Yingjie},
  journal={Annals of Data Science},
  volume={2},
  number={2},
  pages={165--193},
  year={2015},
  publisher={Springer}
}

@article{FLEISCHHACKER20191092,
title = "Portfolio optimization of energy communities to meet reductions in costs and emissions",
journal = "Energy",
volume = "173",
pages = "1092 - 1105",
year = "2019",
issn = "0360-5442",
doi = "https://doi.org/10.1016/j.energy.2019.02.104",
url = "http://www.sciencedirect.com/science/article/pii/S0360544219303032",
author = "Andreas Fleischhacker and Georg Lettner and Daniel Schwabeneder and Hans Auer"
}

@article{kriegel2011density,
  title={Density-based clustering},
  author={Kriegel, Hans-Peter and Kr{\"o}ger, Peer and Sander, J{\"o}rg and Zimek, Arthur},
  journal={Wiley Interdisciplinary Reviews: Data Mining and Knowledge Discovery},
  volume={1},
  number={3},
  pages={231--240},
  year={2011},
  publisher={Wiley Online Library}
}

@article{LI2020123115,
title = "A data-driven strategy to forecast next-day electricity usage and peak electricity demand of a building portfolio using cluster analysis, Cubist regression models and Particle Swarm Optimization",
journal = "Journal of Cleaner Production",
volume = "273",
pages = "123115",
year = "2020",
issn = "0959-6526",
doi = "https://doi.org/10.1016/j.jclepro.2020.123115",
url = "http://www.sciencedirect.com/science/article/pii/S0959652620331607",
author = "Kehua Li and Zhenjun Ma and Duane Robinson and Wenye Lin and Zhixiong Li"
}

@article{MARQUANT201873,
title = "A new combined clustering method to Analyse the potential of district heating networks at large-scale",
journal = "Energy",
volume = "156",
pages = "73 - 83",
year = "2018",
issn = "0360-5442",
doi = "https://doi.org/10.1016/j.energy.2018.05.027",
url = "http://www.sciencedirect.com/science/article/pii/S0360544218308478",
author = "Julien F. Marquant and L. Andrew Bollinger and Ralph Evins and Jan Carmeliet"
}

@article{ahat2013smart,
	title = {Smart {Grid} and {Optimization}},
	volume = {03},
	issn = {2160-8830, 2160-8849},
	url = {http://www.scirp.org/journal/doi.aspx?DOI=10.4236/ajor.2013.31A019},
	doi = {10.4236/ajor.2013.31A019},
	language = {en},
	number = {01},
	urldate = {2020-03-12},
	journal = {American Journal of Operations Research},
	author = {Ahat, Murat and Amor, Soufian Ben and Bui, Marc and Bui, Alain and Guérard, Guillaume and Petermann, Coralie},
	year = {2013},
	pages = {196--206}
}

@Article{Bosom2018multi,
title = {Multi-agent Architecture of a MIBES for Smart Energy Management},
author = {J. {Bosom} and A. {Scius-Bertrand} and H. {Tran} and M. {Bui}},
journal = {Innovations for Community Services. I4CS 2018},
page ={18-32},
volume = {863},
year = {2018}
}

@INPROCEEDINGS{7298002,  author={G. {Guérard} and S. {Ben Amor} and A. {Bui}},  booktitle={2015 International Conference on Smart Cities and Green ICT Systems (SMARTGREENS)},   title={A context-free smart grid model using pretopologic structure},   year={2015},  volume={},  number={},  pages={1-7},  doi={}}

@article{WANG2020117195,
title = "Spatial disparity and hierarchical cluster analysis of final energy consumption in China",
journal = "Energy",
volume = "197",
pages = "117195",
year = "2020",
issn = "0360-5442",
doi = "https://doi.org/10.1016/j.energy.2020.117195",
author = "Shaobin Wang and Haimeng Liu and Haixia Pu and Hao Yang"
}

@article{LI2019735,
title = "An agglomerative hierarchical clustering-based strategy using Shared Nearest Neighbours and multiple dissimilarity measures to identify typical daily electricity usage profiles of university library buildings",
journal = "Energy",
volume = "174",
pages = "735 - 748",
year = "2019",
issn = "0360-5442",
doi = "https://doi.org/10.1016/j.energy.2019.03.003",
url = "http://www.sciencedirect.com/science/article/pii/S0360544219304074",
author = "Kehua Li and Rebecca Jing Yang and Duane Robinson and Jun Ma and Zhenjun Ma"
}

@article{LU201949,
title = "GMM clustering for heating load patterns in-depth identification and prediction model accuracy improvement of district heating system",
journal = "Energy and Buildings",
volume = "190",
pages = "49 - 60",
year = "2019",
issn = "0378-7788",
doi = "https://doi.org/10.1016/j.enbuild.2019.02.014",
url = "http://www.sciencedirect.com/science/article/pii/S0378778818308326",
author = "Yakai Lu and Zhe Tian and Peng Peng and Jide Niu and Wancheng Li and Hejia Zhang"
}

@inproceedings{habib2015outliers,
  title={Outliers detection method using clustering in buildings data},
  author={Habib, Usman and Zucker, Gerhard and Blochle, Max and Judex, Florian and Haase, Jan},
  booktitle={IECON 2015-41st Annual Conference of the IEEE Industrial Electronics Society},
  pages={000694--000700},
  year={2015},
  organization={IEEE}
}

@article{Auray2009Pretopology,
author = {Auray, Jean-Paul and Bonnevay, Stephane and Bui, Marc and Duru, Gérard and Lamure, Michel},
year = {2009},
month = {01},
pages = {27-44},
title = {Pr{\'e}topologie et applications : un {\'e}tat de l'art},
volume = {7},
journal = {Studia Informatica Universalis (Hermann)}
}

@PhDThesis{laborde2019Pretopology,
title = {Pretopology, a mathematical tool for structuring complex systems: methods, algorithms and applications},
author = {Julio {Laborde}},
school = {EPHE},
year = {2019}
}

@phdthesis{le2007classification,
  title={Classification pr{\'e}topologique des donn{\'e}es: application {\`a} l'analyse des trajectoires patients},
  author={Le, Thanh Van},
  year={2007},
  school={Lyon 1}
}

@book{belmandt1993manuel,
  title={Manuel de pr{\'e}topologie et ses applications},
  author={Belmandt, Z},
  year={1993},
  publisher={Herm{\`e}s science publications}
}

@article{largeron2002pretopological,
  title={A pretopological approach for structural analysis},
  author={Largeron, Christine and Bonnevay, St{\'e}phane},
  journal={Information Sciences},
  volume={144},
  number={1-4},
  pages={169--185},
  year={2002},
  publisher={Elsevier}
}

\end{document}